\documentclass[11pt]{article}
\usepackage{fullpage}

\input{irom.sty}

\usepackage[bookmarks=true]{hyperref}
\usepackage{amsthm}
\usepackage{algorithmic}
\usepackage{algorithm}
\usepackage{booktabs}
\usepackage{authblk}
\usepackage{breakcites}
\usepackage[numbers]{natbib}
\usepackage{wrapfig}

\theoremstyle{nospace} 
\theoremstyle{nospace} 
\theoremstyle{nospace} 
\theoremstyle{nospace} 
\theoremstyle{nospace} 
\theoremstyle{nospace} \newtheorem{definition}{Definition}
\theoremstyle{nospace} 
\theoremstyle{nospace} 

\title{Invariant Policy Optimization: Towards Stronger Generalization \\ in Reinforcement Learning}

\author[1]{Anoopkumar Sonar}
\author[2]{Vincent Pacelli}
\author[3]{Anirudha Majumdar}
\affil[1]{Computer Science Department}
\affil[2,3]{Department of Mechanical and Aerospace Engineering}
\affil[ ]{Princeton University}
\affil[ ]{Princeton, NJ, 08544, USA}
\affil[ ]{Emails: \{asonar, vpacelli, ani.majumdar\}@princeton.edu}

\date{}

\begin{document}

\maketitle

\vspace{-20pt}
\begin{abstract}
    A fundamental challenge in reinforcement learning is to learn policies that generalize beyond the operating domains experienced during training. In this paper, we approach this challenge through the following invariance principle: an agent must find a \emph{representation} such that there  exists an  \emph{action-predictor} built on top of this representation that is simultaneously optimal across all training domains. Intuitively, the resulting invariant policy enhances generalization by finding \emph{causes of successful actions}. We propose a novel learning algorithm, Invariant Policy Optimization (IPO), that implements this principle and learns an invariant policy during training. We compare our approach with standard policy gradient methods and demonstrate significant improvements in generalization performance on unseen domains for linear quadratic regulator and grid-world problems, and an example where a robot must learn to open doors with varying physical properties. 
\end{abstract}

%\begin{keywords}%
%  Reinforcement Learning, Generalization, Invariance, Causality
%\end{keywords}

% Sections
\section{Introduction}
\label{sec:intro}

One of the fundamental challenges with state-of-the-art reinforcement learning (RL) approaches is their limited ability to generalize beyond the specific domains they were trained on. The problem of generalization is particularly acute in complex robotics applications. Deploying an RL policy on a robot outside of the laboratory requires learning a policy that can generalize to a wide range of operating domains, especially in safety-critical applications. For example, autonomous vehicles must contend with unfamiliar obstacles, lighting, and road conditions when deployed at scale; robotic manipulators deployed in homes must deal with new objects and environments; and robots operating in close proximity to humans must be able to handle new patterns of human motion.

\begin{figure}[t]
 \begin{center}
 \includegraphics[width=0.8\columnwidth]{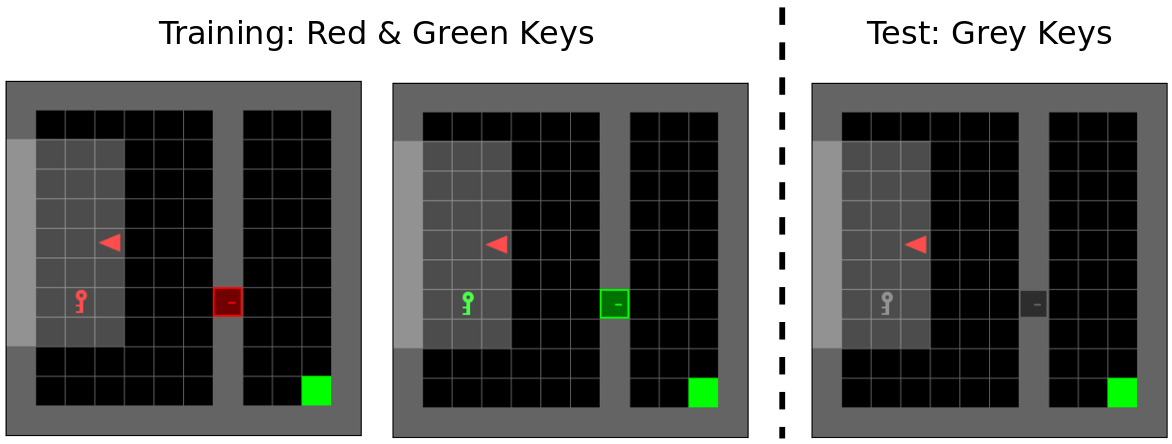}
 \end{center} 
 \vspace{-15pt}
 \caption{\footnotesize{A depiction of the Colored-Key problem described in Section 1 on a $12\times12$ grid. The color of the keys in the environment corresponds to a different operating \emph{domain}. The agent (red triangle) must learn to use the key to open the door and reach the goal (green square). The agent is trained on domains with red and green keys. At test time, the learned policy is deployed on a domain with differently-colored keys (e.g., grey keys). Our results in Section 5 demonstrate that our algorithm generalizes to this novel testing domain significantly better than one trained using standard techniques.} \label{fig:anchor}}
 % \vspace{-15pt}
 \end{figure}

As a simple example, consider the problem shown in Figure \ref{fig:anchor}. A robot is placed in a grid-world and must learn to navigate to a goal located in a different room. In order to do this, it must learn to first navigate to a key, use this key to open the door, and then navigate to the goal. During training, the robot is presented with environments containing red and green keys. A policy trained using standard RL techniques demonstrates strong performance when deployed in test environments with key colors seen during training. However, its performance significantly degrades in test environments with different key colors (see Section \ref{sec:colored keys} for a thorough exploration of this problem). 

Learning policies capable of such generalization remains challenging for a number of reasons. Primarily, RL algorithms have a tendency to memorize solutions to training environments, thereby achieving high training rewards with a brittle policy that will not generalize to novel environments. Moreover, learned policies often fail to ignore distractors in their sensor observations (e.g., the key colors) and are highly sensitive to changes in such irrelevant factors. 
% Moreover, from the design perspective, it can be difficult to predict \emph{a priori} the range of operating domains an agent may encounter and represent them appropriately during training. 
The goal of this paper is to address these challenges and learn policies that achieve strong generalization across new operating domains given a limited set of training domains.

 \begin{figure}[t]
 \centering
 \subfigure[]
 {
 \includegraphics[width=0.276\textwidth]{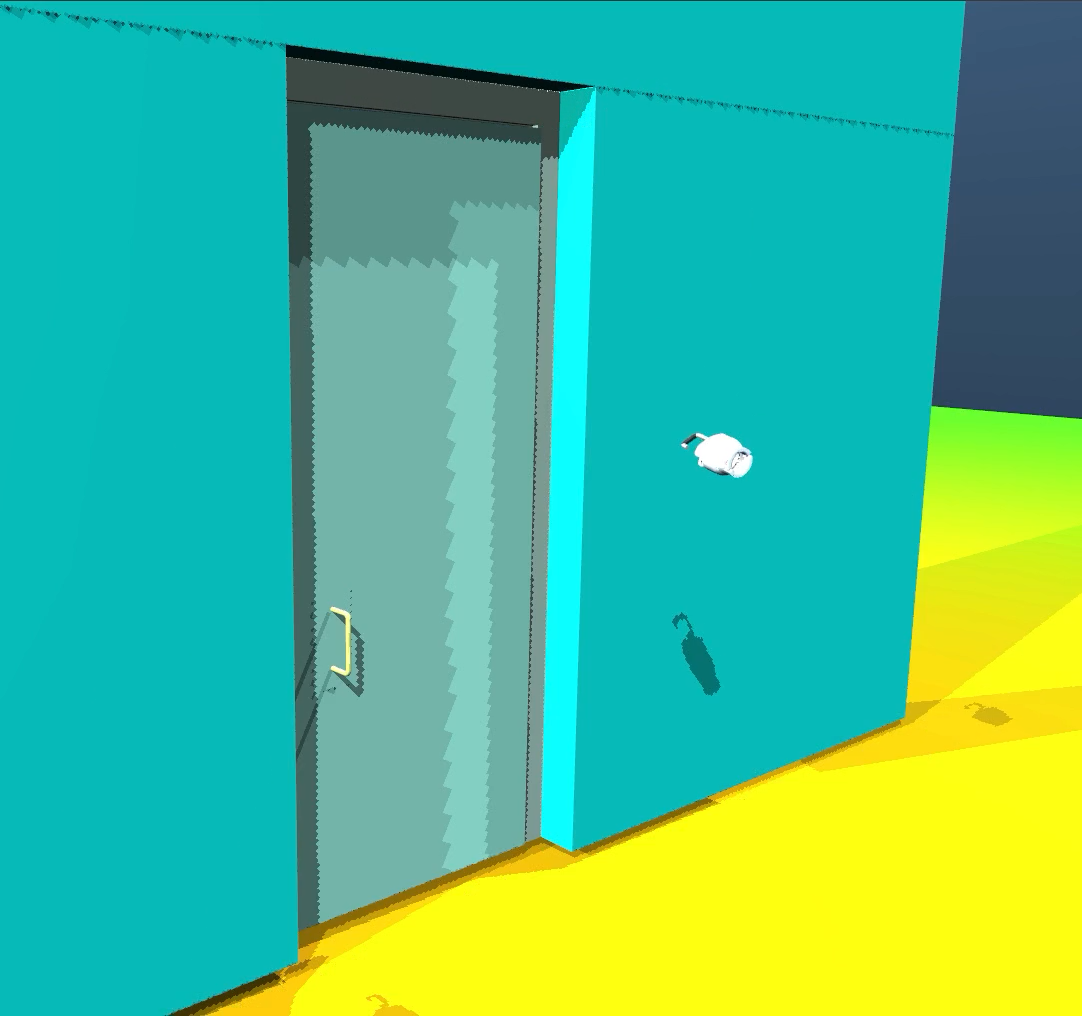}
 \label{fig:dynamics}
 }
 \centering
 % \hspace{10mm}
 \subfigure[]
 {
 \includegraphics[width=0.323\textwidth]{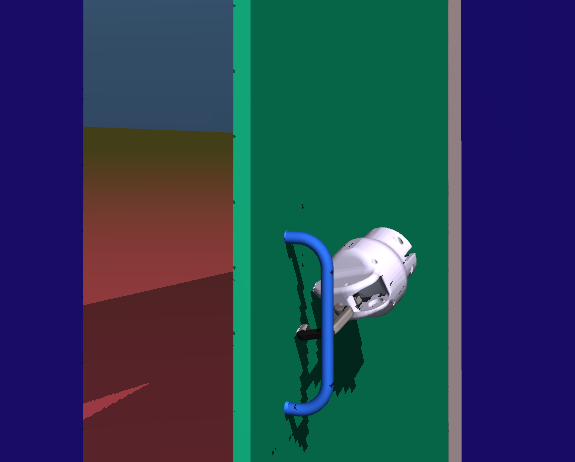}
 \label{fig:policies}
 }
 \subfigure[]
 {
 \includegraphics[width=0.335\textwidth]{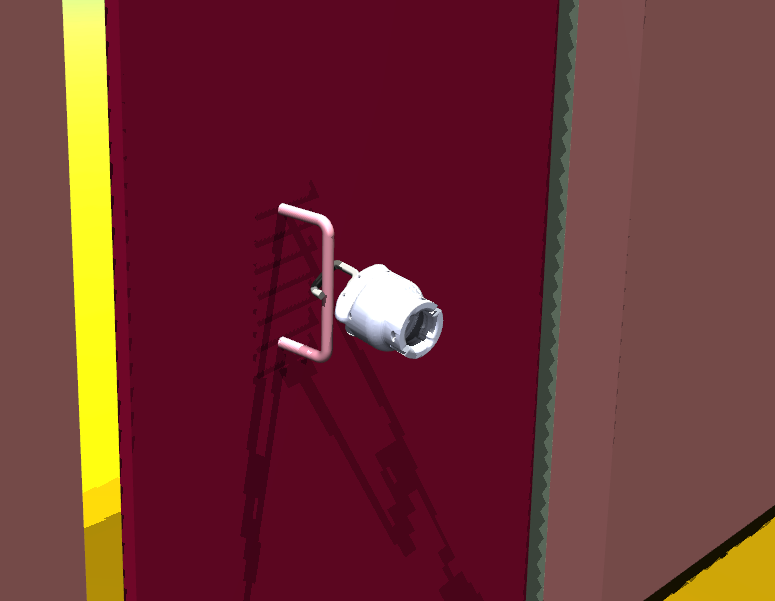}
 \label{fig:policies}
 }
 \vspace{-10pt}
 \caption{\footnotesize{(a) Door-opening environment in DoorGym. Proximal Policy Optimization (PPO) and IPO tend to find \emph{qualitatively} different policies when trained on domains with low door hinge friction. PPO tends to find a policy that uses the outside of the hook (b), while IPO finds a robust hooking strategy (c).} \label{fig:doorgym}}
\vspace{-5pt}
 \end{figure}

\textbf{Statement of Contributions.} We approach the problem of generalizing across domains (formalized in Section 2) with the following principle: a policy will generalize well if it exploits invariances resulting from causal relationships present across domains (e.g. key color does not cause rewards and thus rewards are invariant to the color). To embody this principle, we leverage a close connection between causality and invariance (Section \ref{sec:background}) in an approach we refer to as \emph{Invariant Policy Optimization} (IPO). The key idea is to learn a representation that makes the optimal policy built on top of this representation invariant across training domains. Effectively, this approach attempts to learn and exploit the \emph{causes of successful actions}. We demonstrate that IPO exhibits significantly stronger generalization compared to traditional on-policy methods in three different scenarios (Section \ref{sec:examples}): a linear-quadratic output feedback problem with distracting observations, an instantiation of the colored-key problem, and an example where a robot must learn to open doors with varying physical properties (Figure \ref{fig:doorgym} (a)). 

\subsection{Related Work}
\label{sec:related work}

{\bf Quantifying generalization.} The problem of finding policies that generalize beyond their training domain is becoming an increasingly popular topic as reinforcement learning continues to mature and a number of recent studies have attempted to quantify and understand the generalization challenge in RL. In \cite{Song19}, the authors quantify the effects of \emph{observational overfitting}, where learned policies are sensitive to irrelevant observational features. Benchmark suites including Sonic \cite{Nichol18} and Atari 2600 games \cite{Mnih13} have also been proposed to quantify generalization. Recently, CoinRun \cite{Cobbe19a} and the broader Procgen Benchmark \cite{Cobbe19b} use procedural generalization of environments at controllable levels of difficulty to demonstrate that effective generalization can require an extremely large number of training environments. Another manifestation of the generalization gap is the sim-to-real problem in robotics: agents trained in simulation overfit to this domain and fail to operate in hardware \cite{Tobin17, Peng18, Tan18}.

{\bf Regularization and domain randomization.} The most common approach for improving the out-of-domain generalization of a learning algorithm is to add different forms of regularization. Popular ones borrowed from supervised learning include $L_2$ regularization, dropout \cite{Srivastava14}, and batch normalization \cite{Ioffe15}; each of these has been shown to improve generalization \cite{Cobbe19a}. While practical and easy to implement, these methods typically do not explicitly exploit any structure of the RL problem. Another approach is to constrain the agent's policy to only depend on a set of learned task-relevant variables, which are found by introducing an information-theoretic regularizer \cite{Pacelli20, Goyal19}. These methods have been shown to generalize to new domains that contain task-irrelevant features not present during training. However, the task-relevant variables are not guaranteed to exploit causal relationships in the environment, which is the focus of this paper. Data augmentation and domain randomization have also been shown to be particularly useful in crossing the sim-to-real barrier \cite{Urakami19, Peng18, Akkaya19}. These methods are complementary to the approach presented here and could potentially be used to generate a diverse set of training domains for our method. 

{\bf Distributional robustness.} The PAC-Bayes Control approach \cite{Majumdar20, Veer20} provides a way to make provable generalization guarantees under distributional shifts. This approach is particularly useful in safety-critical applications where it is important to quantify the impact of switching between training and test domains. Another approach that provides robustness guarantees is to train with adversarial perturbations to the underlying data distribution \cite{Sinha17}. However, the challenge with both of these approaches is that they require an \emph{a priori} bound on how much the test domain differs from the training domain (e.g., in terms of an $f$-divergence). In contrast, the recently proposed risk-extrapolation method \cite{Krueger20} promotes out-of-distribution generalization by encouraging robustness of hypotheses over affine combinations of training risks. This method is shown to improve performance of RL agents when their state space is augmented with noisy copies of true system states.

{\bf Causality and invariance.} Recently, the task of learning causal predictors has drawn interest in the supervised learning setting. An approach formalized in \cite{Peters16} attempts to find features that are causally linked to a target variable by exploiting the invariance of causal relationships \cite{Pearl09, Peters17}. This idea was expanded upon in the invariant risk minimization (IRM) approach \cite{Arjovsky19}, which formulates the problem in terms of finding a representation such that the optimal classifier built on top of this representation is invariant across domains. This results in classifiers that ignore spurious correlations that may exist in any single domain. The formulation leads to a challenging bilevel optimization problem and is tackled via a regularizer that approximates its solution.
In \cite{Ahuja20}, the authors present a game-theoretic reformulation of the IRM principle and propose a new algorithm, known as IRM-Games, which offers better empirical results. The approach presented in \cite{Teney20} uses a variance regularization scheme to find approximately-invariant classifiers. Our approach adapts ideas from causality and invariance to RL settings by learning representations that invariantly predict actions across domains. We provide more background on invariance, causality, and IRM in Section \ref{sec:background}. 

{\bf Causality in RL.} Lastly, there are a number of recent methods that attempt to exploit causality in RL. For example, \cite{Dasgupta19} observed that, in some instances, causal reasoning can emerge in agents trained via meta-learning. Other approaches explicitly attempt to learn causal graphs that describe the dynamics of the agent's environment. Along these lines, \cite{Nair19} proposes a two-phase training process where interactions with the environment are first used to learn the causal graph of the environment, and then a policy that exploits this graph is trained. Finally, the IRM method has recently been applied to RL problems. In \cite{Zhang20}, the authors attempt to learn a causal Markov decision process (MDP) that is bisimilar to the full MDP present during training. This formulation requires learning a model for both the causal and full dynamics of the system, a mapping between the two, and a causal model of the rewards. Standard RL algorithms are then used in conjunction with these causal models to produce a final policy. This approach is distinct from the one in this paper, which does not seek to find complete dynamical models (causal or otherwise). Instead, we focus on \emph{identifying the causes of successful actions}, which is potentially significantly simpler.

\section{Problem Formulation}
\label{sec:problem formulation}

We are interested in the problem of zero-shot generalization to environments that can be significantly different from environments seen during training. We formalize this as follows. Let $s_t = (s_t^a, s_t^e)$ denote the joint state of the agent and environment. In the colored-keys example (Section \ref{sec:intro}), $s_t^a$ corresponds to the location of the robot at time $t$, while $s_t^e$ corresponds to locations of the obstacles, key, door, and goal. In our formulation, different environments correspond to different (initial) states of the environment (e.g., different configurations of obstacles, key, door, and goal). We denote the agent's actions, observations, and rewards by $a_t \in \mathcal{A}$, $o_t \in \mathcal{O}$, and $r_t \in \mathbb{R}$ respectively. 

During training, we assume access to multiple sets of environments $\{e_i^d\}_{i=1}^{n_d}$ from different \emph{domains} $d \in D_{\text{tr}}$. We assume that the action space $\mathcal{A}$ and observation space $\mathcal{O}$ are shared across all domains (state spaces need not be shared). Each domain $d$ corresponds to a partially observable Markov decision process (POMDP) \cite{Thrun05} with dynamics mapping $p^d(s_{t+1} | s_t, a_t)$, observation mapping $p^d(o_t | s_t)$, and reward mapping $p^d(r_t | s_t, a_t)$. In the colored-keys example, domains differ (only) in terms of the observation mapping; in particular, each domain assigns a particular color to keys. Each domain also defines a distribution $\mathcal{D}^d$ over environments.  

Our goal is to learn a policy that generalizes to domains $D_\text{all} \supseteq D_\text{tr}$ \emph{beyond} the training domains (e.g., generalizing to key colors not seen during training). Specifically, let $R_e(\pi)$ denote the expected cumulative reward $\sum_t r_t$ (over a finite or infinite horizon) when policy $\pi$ is executed in environment $e$. We would then like to maximize the worst-case rewards over all domains:   
\begin{equation}
    R^\text{all}(\pi) := \underset{d \in D_\text{all}}{\min} R^d(\pi) := \underset{d \in D_\text{all}}{\min} \ \underset{e \sim  \mathcal{D}^d}{\mathbb{E}} \big{[}R_e(\pi)\big{]}.
\end{equation}
Without further assumptions on the relationship between $D_\text{tr}$ and $D_\text{all}$, finding a policy that performs well on domains $D_\text{all}$ may be impossible. We discuss this further in Section \ref{sec:approach}.

\section{Background: Invariance and Causality}
\label{sec:background}

In this section, we provide a brief exposition of causality and its relationship to invariance. We refer the reader to \cite{Peters17, Arjovsky19, Pearl09, Ahuja20} for a more thorough introduction. % We start by defining a structural causal model (also known as a causal graph), present the causal graph for our RL problem formulation and then define the notion of an intervention on a causal graph. Consequently, we invoke the modularity principle from causality theory which allows us to use invariance as a proxy into the underlying causation and helps find causal relationships. The modularity principle further allows us to treat each domain during training as an intervention on the RL causal graph. We then summarize two existing techniques, Invariant Risk Minimization (IRM) and IRM-Games that enable us to find invariances, and hence causes of successful actions, across domains.\\ \ani{This seems redundant to me. I would suggest reverting it back to the original version.}

\begin{definition}[Structural Causal Model \cite{Peters17}]
A \emph{structural causal model (SCM)} $\mathcal{C} = (S, \eta)$ governing the random vector $x := (x_1, \dots, x_m)$  is a collection $S$ of $m$ assignments:
\begin{equation}
    S_j: x_j \leftarrow f_j(\text{\emph{Pa}}(x_j), \eta_j), \ \ j = 1, \dots, m,
\end{equation}
where $\text{\emph{Pa}}(x_j) \subseteq \{x_1, \dots, x_m\} \backslash \{x_j\}$ are the \emph{parents} of $x_j$, the $\eta_1, \dots, \eta_m$ are independent noise variables, and each $f_j$ is a mapping from these variables to $x_j$. The graph $\mathcal{G}$ of an SCM is obtained by associating one vertex for each $x_j$ and edges from each parent in $\text{\emph{Pa}}(x_j)$ to $x_j$. We assume acyclic causal graphs. We refer to the elements of $\text{\emph{Pa}}(x_j)$ as the \emph{direct causes} of $x_j$.
\end{definition}

% \begin{figure}[t]
% \centering
% \subfigure[]
% {
% \includegraphics[width=0.5\textwidth]{sections/figures/dynamics_causal_graph.pdf}
% \label{fig:dynamics}
% }
% \centering
% \hspace{10mm}
% \subfigure[]
% {
% \includegraphics[width=0.09\textwidth]{sections/figures/dynamics_causal_graph.pdf}
% \label{fig:policies}
% }
% \vspace{1mm}
% \caption{(a) Causal graph corresponding to the RL setting we consider. Here, $s_t$ is the state, $o_t$ is the observation, and $a_t$ is the action. The reward $r_t$ depends only on $\tilde{s}_t$ and $a_t$. (b) We seek a representation $\Phi$ such that there exists $\pi$ that is simultaneously optimal across domains.}
% % \vspace{-5mm}
% \end{figure}

\begin{definition}[Intervention \cite{Peters17}]
Consider an SCM $\mathcal{C} = (S, \eta)$. An \emph{intervention} $d$ changes one or more of the structural assignments to obtain a new SCM $\mathcal{C}^d = (S^d, \eta^d)$ with assignments:
\begin{equation}
    S_j^d: x_j \leftarrow f_j^d(\text{\emph{Pa}}^d(x_j), \eta_j^d), \ \ j = 1, \dots, m.
\end{equation}
We say that the variables whose structural assignments we have changed have been \emph{intervened on}.
\end{definition}
\begin{wrapfigure}{R}{0.45\textwidth}
\vspace{-15pt}
\centering
   \subfigure[]
   {\includegraphics[width=0.3\textwidth]{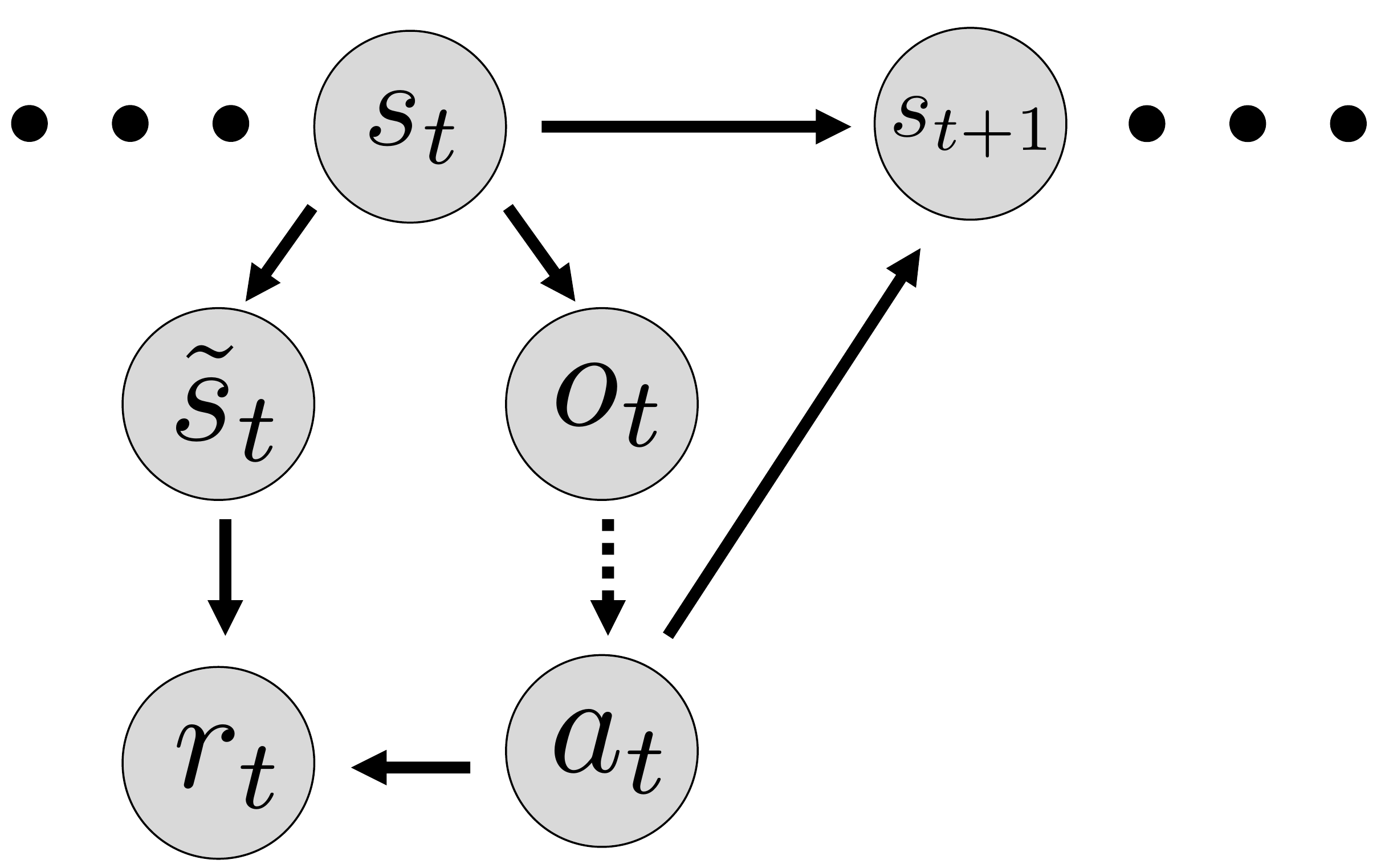} \label{fig:dynamics}}
   \hfill
   \subfigure[]
   {\includegraphics[width=0.07\textwidth]{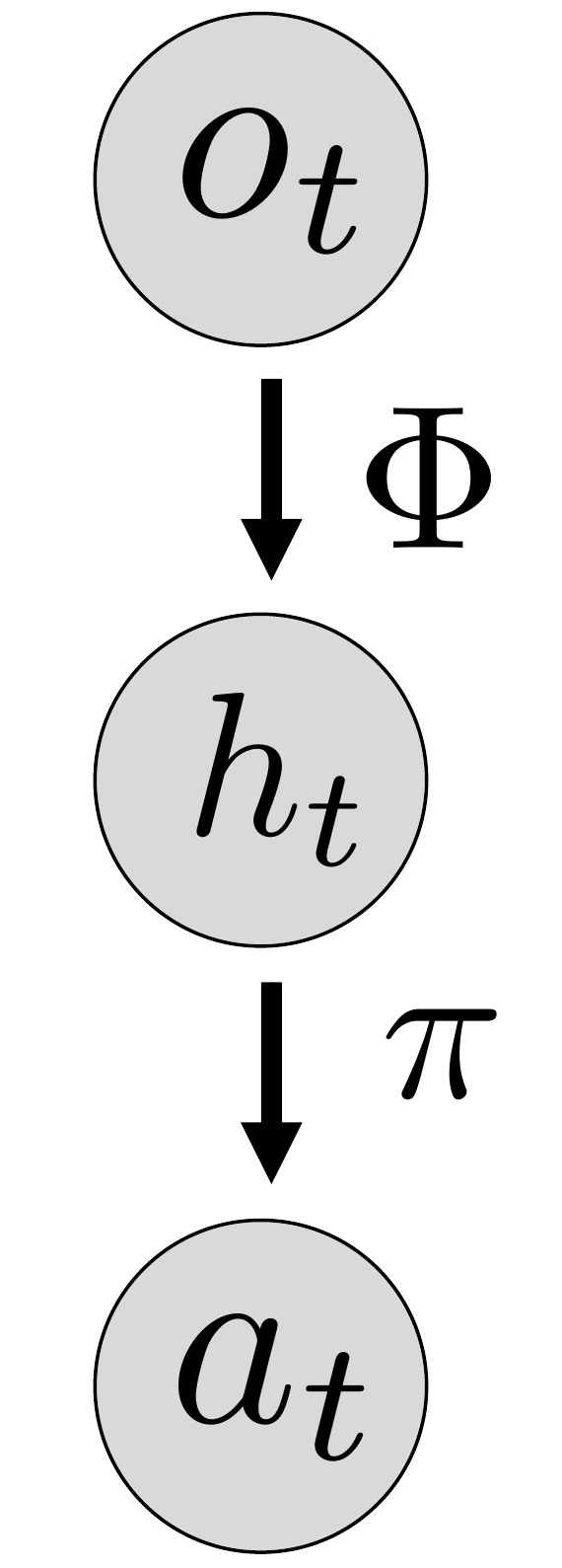} \label{fig:policies}}
   \vspace{-10pt}
\caption{\footnotesize{(a) Causal graph corresponding to our RL setting. Here, $s_t$ is the state, $o_t$ is the observation, and $a_t$ is the action. The reward $r_t$ depends only on $\tilde{s}_t$ and $a_t$. (b) We seek a representation $\Phi: o_t \mapsto h_t$ such that there exists $\pi: h_t \mapsto a_t$ that is simultaneously optimal across domains.} \label{fig:diagrams}}
\vspace{-5pt}
\end{wrapfigure} 
Figure \ref{fig:diagrams}(a) shows the causal graph for the RL formulation in Section \ref{sec:problem formulation}. Here, the reward $r_t$ depends on the action $a_t$ and a set of ``reward-relevant" variables $\tilde{s}_t$. Thus, $\text{Pa}(r_t) = \{\tilde{s}_t\} \cup \{a_t\}$. In our running colored-keys example, $\tilde{s}_t$ is purely a function of the agent's state $s_t^a$ and the goal location. 

\textbf{Modularity principle} \cite{Bareinboim12, Peters17}. The modularity principle establishes a close relationship between causality and \emph{invariance}: a set of variables $\{x_1, \dots, x_k\}$ are the direct causes of $y$ if and only if the conditional probability $p(y | x_1, \dots, x_k)$ remains invariant for all interventions where $y$ has not been intervened on. This is also related to the notion of ``autonomy"  and the principle of independent mechanisms \cite[Ch. 2.1]{Peters17}. As an example, consider the reward $r_t$ to be the variable of interest in Figure \ref{fig:dynamics}. Then, $(\tilde{s}_t, a_t)$ are the direct causes of $r_t$ if and only if for all interventions where $r_t$ has not been intervened on, $p(r_t | \tilde{s}_t, a_t)$ remains invariant. Thus, in the context of the colored-keys example, $\tilde{s}_t$ does not contain any color-related information. 

\textbf{Invariant Risk Minimization (IRM).} IRM \cite{Arjovsky19} exploits the modularity principle in the context of supervised learning. One assumes datasets $\{(x_i^d, y_i^d)\}_{i=1}^{n_d}$ from multiple training domains\footnote{We note that \cite{Arjovsky19} uses the term ``environment" instead of ``domain". However, we use ``domain" since ``environment" has a different meaning in RL contexts.} $d \in D_\text{tr}$ corresponding to different interventions on the data-generating process that do not intervene on the target variable $y$. Here $x_i^d \in \mathcal{X} \subseteq \mathbb{R}^{n_x}$ and $y_i^d \in \mathcal{Y} \subseteq \mathbb{R}^{n_y}$. The goal is to learn a data representation $\Phi: \mathcal{X} \rightarrow \mathcal{H} \subseteq \mathbb{R}^{n_h}$ that elicits an \emph{invariant predictor} $w \circ \mathcal{H}$ across training domains, i.e., a representation $\Phi$ such that there exists a classifier $w: \mathcal{H} \rightarrow \mathcal{Y}$ that is \emph{simultaneously optimal} for all training domains $d \in D_\text{tr}$. Intuitively, the representation $\Phi$ should capture the direct causes of $y$ and thus eliminate any features in $x$ that spuriously correlate with $y$. 
The optimization problem associated with IRM is a challenging bi-leveled one. The authors of \cite{Arjovsky19} propose IRM-v1, where a regularization scheme is used to find a representation $\Phi$ that leads to classifiers that are approximately locally optimal across training domains. 

% one learns a representation $\Phi: \mathcal{X} \rightarrow \mathcal{Y}$ that is approximately locally optimal in all training domains. % fixes a ``dummy" linear classifier $w = 1.0$ and learns a representation $\Phi: \mathcal{X} \rightarrow \mathcal{Y}$ that is approximately locally optimal in all training domains:
% \begin{equation}
%     \underset{\Phi: \mathcal{X} \rightarrow \mathcal{Y}}{\min}  \sum_{d \in D_\text{tr}} L^d(\Phi) + \lambda \cdot \|\nabla_{w|w=1.0} \  L^d(w \circ \Phi) \|^2,
% \end{equation}
% where $L^d(\Phi)$ is the loss incurred by $\Phi$ on domain $d$. 

\textbf{IRM Games.} Inspired by IRM, the authors of \cite{Ahuja20} demonstrate that the set of invariant predictors corresponds to the set of pure Nash equilibria of a game played among $n_d := |D_\text{tr}|$ players. Each player (corresponding to a training domain $d$) can choose its own classifier $w^d$ and is trying to maximize its own utility function: $-L^d(w^\text{av} \circ \Phi)$, where $L^d(\cdot)$ is the loss on domain $d$ and $w^\text{av}(\cdot) := \frac{1}{n_d} \sum_{d=1}^{n_d} w^d(\cdot)$. While finding Nash equilibria for continuous games is challenging in general, the game theory literature contains several heuristic schemes. In \cite{Ahuja20}, the authors propose a strategy based on \emph{best response dynamics} \cite{Barron10}, where players take turns maximizing their utility functions. The resulting algorithm achieves similar or better empirical performance as compared to IRM-v1, with significantly reduced variance.

\section{Invariant Policy Optimization}
\label{sec:approach}

We now describe our novel reinforcement learning algorithm, which we refer to as \emph{invariant policy optimization} (IPO). The key insight behind this algorithm is to implement the following invariance principle: learn a representation $\Phi: \mathcal{O} \mapsto \mathcal{H} \subseteq \mathbb{R}^{n_h}$ that maps observations $o_t$ to $h_t \in \mathcal{H}$ in a manner that supports \emph{invariant action prediction} (see Figure \ref{fig:diagrams}(b)). More precisely, the goal is to learn a representation $\Phi$ such that there exists an ``action-predictor" $\pi: \mathcal{H} \rightarrow \mathcal{A}$ built on top of this representation that is \emph{simultaneously optimal across all training domains}\footnote{For the ease of exposition, we discuss the case of memoryless policies. However, it is straightforward to handle policies with memory (e.g., by augmenting observations with a memory state).}. We will refer to the resulting policy $\pi \circ \Phi$ as an \emph{invariant policy}. This invariance principle can be formally embodied as the following optimization problem:
% \begin{small}
\begin{flalign}
\underset{\begin{subarray}{c}
  \Phi: \mathcal{O} \rightarrow \mathcal{H} \\
  \pi: \mathcal{H} \rightarrow \mathcal{A}
  \end{subarray}}{\text{max}}~~~& \sum_{d=1}^{n_d} R^d(\pi \circ \Phi) \label{opt:ipo-actions-orig} \\
\text{s.t.}~~~& \pi \in \underset{\bar{\pi}: \mathcal{H} \rightarrow \mathcal{A}}{\text{argmax}} \ \ R^d\Big{(}\bar{\pi} \circ \Phi\Big{)}, \ \ \forall d = 1, \dots, n_d. \nonumber
\end{flalign}
% \end{small}
Here, $R^d$ is the reward associated with domain $d$, as defined in Section \ref{sec:problem formulation}. 
Intuitively, given a set of training domains, IPO attempts to learn a representation that corresponds to the ``causes of successful actions". This interpretation elucidates the role of the different training domains; these must correspond to different interventions on the causal graph shown in Figure \ref{fig:dynamics} that leave optimal actions unaffected. Assuming a diverse set of training domains, one learns a representation that eliminates features that spuriously correlate with good actions (i.e., actions that achieve high rewards). For example, in the colored-keys problem, such a representation corresponds to one that eliminates color from observations. By eliminating such features, an invariant policy generalizes well to novel domains corresponding to unseen interventions on the spurious/irrelevant features.  % We note that a representation that invariantly predicts actions may exist even when a representation that invariantly predicts rewards does not exist. For example, consider a problem with two domains where the reward functions differ in a way that allows for a shared optimal policy (e.g., reward functions that differ by a constant offset).\anoop{I don't completely understand this idea. Perhaps elaborating a little on the example might help}

Our algorithmic approach for IPO is inspired by the game-theoretic formulation of \cite{Ahuja20} (see Section \ref{sec:background}). We endow each domain $d$ with its own policy $\pi^d$ and define an overall averaged policy $\pi^\text{av}(h_t) := \frac{1}{n_d} \sum_{d=1}^{n_d} \pi^d(h_t)$. Further details on this averaging step for stochastic policies are provided in Appendix \ref{sec:averaging}.
The optimization problem behind IPO then becomes:
% \begin{small}
\begin{flalign}
\underset{\Phi, \pi^1,\dots,\pi^d}{\text{max}}~& \sum_{d=1}^{n_d} R^d \Bigg{(} \underbrace{\frac{1}{n_d} \sum_{d=1}^{n_d} \pi^d \circ \Phi}_{:= \pi^\text{av} \circ \Phi} \Bigg{)} \label{opt:ipo-actions} \\
\text{s.t.}~~~& \pi^d \in \underset{\bar{\pi}^d}{\text{argmax}} \ \ R^d\Bigg{(}\frac{1}{n_d} \Big{[}\bar{\pi}^d \circ \Phi + \sum_{i \neq d} \pi^i \circ \Phi \Big{]} \Bigg{)}, \ \ \forall d = 1, \dots, n_d. \nonumber
\end{flalign}
% \end{small}
Next, we relate Problem \eqref{opt:ipo-actions} to a game played between $n_d$ players. Each player corresponds to a domain $d$ and chooses a policy $\pi^d$ to maximize its own utility function $R^d(\pi^\text{av} \circ \Phi)$. Since Problem \eqref{opt:ipo-actions} is identical to the one in \cite{Ahuja20} for finding invariant representations (with policies playing the role of classifiers), the results from \cite{Ahuja20} carry over to our setting. In particular, under mild technical assumptions on the policies, the set of pure Nash equilibria of the game correspond to the set of invariant policies. We refer the reader to \cite{Ahuja20} for details on the technical assumptions, but note that these are satisfied by a wide range of function classes (e.g., ReLu networks with arbitrary depth, linear functions, and functions in $L^p$ spaces). 

While finding Nash equilibria for continuous games such as the one above is difficult in general, the game theory literature has developed several approximate approaches that demonstrate good performance in practice. Here, we adapt the strategy based on \emph{best response dynamics} \cite{Barron10} proposed in \cite{Ahuja20} to our setting. The resulting IPO training procedure is presented in Algorithm \ref{a:ipo}. The for-loop in lines 8--11 implement the best-response dynamics; the players (corresponding to the different domains) take turns choosing $\pi^d$ in order to optimize their own objective $R^d(\pi^\text{av} \circ \Phi)$. We choose to implement the updates using proximal policy optimization (PPO) \cite{Schulman17}. However, this choice is not fundamental and one may implement the updates using other policy gradient methods. Algorithm \ref{a:ipo} can also accommodate actor-critic methods. In this version, each domain $d$ has both an actor $\pi^d$ and a critic $v^d$. In the policy-update steps, one updates both the actor and the critic using PPO. 

Line 5 of the algorithm periodically updates the representation $\Phi$. 
However, as demonstrated in \cite{Ahuja20}, simply fixing $\Phi$ = I (identity) is an effective approach and can (under certain conditions) recover invariant predictors that extract causal features (and ignore non-causal factors such as color in the colored-key example). The intuition is that the burden of extracting causal features from observations is simply shifted to $\pi$ (the portion of the overall policy that maps the output of $\Phi$ to actions). % Concretely, consider a setting where the observation o = [o_c, o_nc] consists of two components corresponding to causal features o_c and non-causal features o_nc (e.g., color). Then, one can construct an invariant policy of the form o -> Phi(o) -> pi(Phi(o)), where Phi extracts the causal features from the observation (i.e., Phi maps o to o_c). Now, suppose one had chosen Phi = Identity for this example. Then, one can still construct an invariant policy: o -> Identity(o) -> pi'(Identity(o)), where pi' is pi(Phi(o)). Hence, the overall policy is still extracting causal features, but the burden of this extraction has simply been transferred to pi'. The paper [23] provides a more formal characterization of conditions under which one may choose Phi = Identity. 
In our numerical experiments (Section \ref{sec:examples}), we did not find significant benefits to periodically updating $\Phi$. Further discussion on the choice of $\Phi$ = I is provided in Appendix \ref{sec:phi}. 

\begin{algorithm}[t] 
	% \algsetup{linenosize=\small} % \normalsize}
  \small
  \caption{{\small Invariant Policy Optimization (IPO)}}
  \label{a:ipo}
  \begin{algorithmic}[1]
   % \STATE Fix $\Phi = $ I (identity map)
    % \STATE Define $\pi^\text{av}(\cdot) := \frac{1}{n_d} \sum_{d=1}^{n_d} \pi^d(\cdot)$
    \FOR{iter = 1, 2, \dots $\text{iter}_\text{max}$} 
    \IF{Fixed-$\Phi$} 
    \STATE $\Phi = $ I (identity map)
    \ELSE 
    \STATE $\Phi \leftarrow \text{PPO} \Big{(}\sum_{d=1}^{n_d} R^d(\pi^\text{av}_\text{cur} \circ \Phi) \Big{)}$ $//$ Update $\Phi$ via an iteration of proximal policy optimization
    \ENDIF
    \FOR{$i = 1, \dots, K$}
    \FOR{$d = 1, \dots, n_d$}
    \STATE $\pi^d \leftarrow \text{PPO} \Big{(} R^d(\pi^\text{av} \circ \Phi) \Big{)}$ $//$ Update $\pi^d$ while keeping $\pi^i$ with $i \neq d$ fixed. 
    \STATE Define $\pi^\text{av}(\cdot) := \frac{1}{n_d} \sum_{d=1}^{n_d} \pi^d(\cdot)$$//$ Averaging
    \ENDFOR
    \ENDFOR
    \ENDFOR
  \end{algorithmic}
\end{algorithm}

%% Older version of algorithm
%\begin{algorithm}[t] 
%	% \algsetup{linenosize=\small} % \normalsize}
%  \small
%  \caption{{\small Invariant Policy Optimization (IPO)}}
%  \label{a:ipo}
%  \begin{algorithmic}[1]
%    \STATE Fix $\Phi = $ I (identity map)
%    \STATE Define $\pi^\text{av}(\cdot) := \frac{1}{n_d} \sum_{d=1}^{n_d} \pi^d(\cdot)$
%    \FOR{iter = 1, 2, \dots $\text{iter}_\text{max}$} 
%    % \IF{Fixed-$\Phi$} 
%    % \STATE $\Phi = $ I
%    % \ELSE 
%    % \STATE $\Phi \leftarrow \text{PPO} \Big{(}\sum_{d=1}^{n_d} R^d(\pi^\text{av}_\text{cur} \circ \Phi) \Big{)}$ $//$ Update $\Phi$ via an iteration of proximal policy optimization
%    % \ENDIF
%    % \FOR{$i = 1, \dots, K$}
%    \FOR{$d = 1, \dots, n_d$}
%    \STATE $\pi^d \leftarrow \text{PPO} \Big{(} R^d(\pi^\text{av} \circ \Phi) \Big{)}$ $//$ Update $\pi^d$ while keeping $\pi^i$ with $i \neq d$ fixed. 
%    % \STATE $\pi^\text{av} \leftarrow \frac{1}{n_d} \sum_{d=1}^{n_d} \pi^d$ $//$ Averaging
%    % \ENDFOR
%    \ENDFOR
%    \ENDFOR
%  \end{algorithmic}
%\end{algorithm}

\section{Examples}
\label{sec:examples}

Next, we demonstrate the generalization benefits afforded by IPO on three examples. 
% : (i) linear quadratic regulator problems with distractor observations, (ii) an instantiation of the colored-keys problem, and (iii) a robotic manipulation example where an agent must learn to open doors with varying physical properties. 
Code for all our examples is available on GitHub: \href{https://github.com/irom-lab/Invariant-Policy-Optimization.git}{github.com/irom-lab/Invariant-Policy-Optimization.git}. 

\subsection{Linear Quadratic Regulator with Distractors}
\label{sec:lqr}

We first apply our approach to the linear quadratic regulator (LQR) problem \cite{Anderson07} modified to include high-dimensional ``distractor" observations. There has been a growing interest in LQR as a simplified surrogate for deep RL problems \cite{Dean19, Tu18, Fazel18, Agarwal19}. Here we consider the output-feedback control problem proposed in \cite{Song19} as a benchmark for assessing generalization with respect to changes in the observation model. The dynamics of the system are described by 
$s_{t+1} = As_t + Ba_t$,
where $s_t \in \mathbb{R}^{n_s}, a_t \in \mathbb{R}^{n_a}$, and $A, B$ are fixed matrices. The agent receives a high-dimensional sensor observation 
$o_t = \begin{bmatrix} W_c \\ W_d \end{bmatrix} s_t$,
where $W_c$ and $W_d$ are semi-orthogonal matrices. This ensures that the portion of the observation corresponding to $W_c s_t \in \mathbb{R}^{n_s}$ contains full information about the state, while $W_d s_t \in \mathbb{R}^{n_y}$ is a high-dimensional ``distractor". The goal is to choose policies of the form $a_t = K o_t$ in order to minimize the infinite-horizon LQR cost $\sum_{t=0}^\infty s_t^T Q s_t + a_t^T R a_t$. % In the case where $o_t = s_t$, one can find the optimal policy via gradient descent even though the corresponding optimization problem is non-convex \cite{}. 

Here, a domain corresponds to a particular choice of $W_d$; all other parameters ($A, B, Q, R, W_c$) are shared across domains and unknown to the agent. During training time, one learns a policy $K o_t$ using $n_d$ domains. %\anoop{We've defined the idea of a domain in the LQR context but is there a similar analog for an environment drawn from the domain? Or is there only one environment corresponding to each domain?} \ani{You could think about each initial state as corresponding to an environment. But, I think this will just cause confusion if we present it like that.}%
At test time, the learned policy is assessed on a new domain. In the case where there is a single domain (used for both training and test) and $o_t = s_t$, one can find the globally optimal policy via gradient descent (despite the non-convexity of the problem) \cite{Fazel18}. However, as demonstrated in \cite{Song19}, simple policy gradient using the combined costs of multiple training domains finds a policy that overfits to the training domains in the more general setting considered here. Intuitively, this is because the learned policy fails to ignore the distractors. 

For our numerical experiments, we choose $n_x = n_a = 20$ and $Q=R=I_{20 \times 20}$. The matrices $A$ and $W_c$ are random orthogonal matrices, $B$ is $I_{20 \times 20}$, and the $W_d$ are random semi-orthogonal matrices (different for each domain). For IPO, we employ a policy $\frac{1}{n_{d}}\sum_{i=1}^{n_d} K_i^d$ that averages policies $K^d$ corresponding to the training domains. Instead of PPO, we simply use gradient descent to perform policy updates. Optimization hyperparameters are provided in Appendix \ref{sec:lqr hyperparams}.  

We compare our approach with two baselines: (i) gradient descent on $K$ using the combined cost of $n_d$ training domains, and (ii) gradient descent using an \emph{overparameterized} class of policies with two layers (i.e., $K = K_1 K_2$) and hidden dimension of $10 n_a$. Interestingly, \cite{Song19} found that this form of overparameterization induces an implicit regularization towards ``simpler" policies (i.e., ones that are less ``dependent" on the distractors). We present results using both the Fixed-$\Phi$ version of IPO (ref. Algorithm \ref{a:ipo}) and the version where $\Phi$ is optimized. In the latter version, we choose $\Phi$ to be linear (with output dimension $10 n_a$). 
 Table \ref{tab:lqr domains} compares the generalization performance of the learned policies to new domains as we vary the number of training domains. Here, the distractors have dimension $1000$. Consistent with \cite{Song19}, we find that overparameterization forms a strong baseline for this problem. However, IPO significantly outperforms both baselines. As expected, performance improves with increasing number of training domains and tends towards the performance achieved by an ``oracle" policy that has access to the full state $s_t$ on the test domain. We note that the results using Fixed-$\Phi$ are largely similar to the version with optimized $\Phi$ (consistent with the arguments provided in Section \ref{sec:approach} and Appendix \ref{sec:phi}). Table \ref{tab:lqr distractor} assesses the impact of changing the distractor dimension. Here, we fix the number of training domains to five. Again, we observe that IPO demonstrates significantly improved performance. 
 
\begin{table}[t]
\footnotesize
\begin{center}
% \scalebox{0.9}{
  \begin{tabular}{ | l | c | c | c | c| c | }
    \hline 
    Number of training domains & 2 & 3 & 4 & 5 & 10 \\ \hline
    \hline
    Gradient descent &  97.7$\pm$5.4  & 90.0$\pm$9.2 & 82.6$\pm$4.3 & 78.6$\pm$5.4 & 68.8$\pm$3.9 \\ \hline
    Overparameterization &  86.2$\pm$3.0  & 75.3$\pm$4.4  & 69.4$\pm$1.8 & 64.5$\pm$1.9 & 51.4$\pm$1.0 \\ \hline
    IPO (ours) [Fixed-$\Phi$] &  78.8$\pm$3.5  & 64.8$\pm$2.3  & {\bf 57.7$\pm$1.0} & {\bf 52.3$\pm$1.8} & {\bf 43.2$\pm$1.1} \\ \hline
    IPO (ours) [Variable-$\Phi$] &  {\bf 71.3$\pm$2.1}  & {\bf 62.7$\pm$2.4}  & 57.8$\pm$1.0 & 53.9$\pm$1.1 & 45.3$\pm$0.45 \\ \hline
    LQR oracle &  32.1  & 32.1  & 32.1 & 32.1 & 32.1 \\
    \hline
  \end{tabular}
 % }
  \vspace{-2pt}
    \caption[]{\footnotesize{LQR with distractors: comparison of IPO with two baselines (gradient descent and overparameterization) with distractor dimension $n_y = 1000$ and varying number of training domains. IPO demonstrates stronger generalization (lower costs) compared to the baselines. The mean and std. dev. are across 10 seeds.}  \label{tab:lqr domains}}
  \end{center}
 % \vspace{-20pt}
  \end{table}

\begin{table}[t]
\footnotesize
\begin{center}
  \begin{tabular}{ | l | c | c | c | c| c | }
    \hline 
    Distractor dimension & 100 & 500 & 1000 & 1500 & 2000 \\ \hline
    \hline
    Gradient descent &  46.3$\pm$2.4  & 70.3$\pm$4.5 & 78.6$\pm$5.4 & 88.7$\pm$5.6 & 94.7$\pm$8.2 \\ \hline
    Overparameterization & \bf 36.3$\pm$0.5  & 54.3$\pm$2.0  & 64.5$\pm$1.9 & 71.6$\pm$2.0 & 79.1$\pm$3.4 \\ \hline
    IPO (ours) [Fixed-$\Phi$] &  42.5$\pm$2.0  & 50.8$\pm$1.1  & {\bf 52.3$\pm$1.8} & {\bf 55.0$\pm$1.7} & {\bf 59.4$\pm$2.2} \\ \hline
IPO (ours) [Variable-$\Phi$] &  {38.5$\pm$0.6}  & {\bf 48.0$\pm$1.1}  & 53.9$\pm$1.1 & 58.1$\pm$1.5 & 62.0$\pm$1.6 \\ \hline
    LQR oracle &  32.1  & 32.1  & 32.1 & 32.1 & 32.1 \\
    \hline
  \end{tabular}
  \vspace{-2pt}
    \caption[]{\footnotesize{LQR with distractors: comparison of IPO with two baselines (gradient descent and overparameterization) with varying dimensionality of distractors and $n_d = 5$. IPO demonstrates stronger generalization (i.e., lower costs) compared to the two baselines. The reported mean and std. dev. are across 10 different seeds.}  \label{tab:lqr distractor}}
  \end{center}
  \vspace{-20pt}
  \end{table}
  
% \vspace{-2pt}
\subsection{Colored-Key Domains}
\label{sec:colored keys}
% \vspace{-2pt}

We now consider the colored-keys problem introduced in Section \ref{sec:intro}. In this example, a robot is placed in a grid-world that contains a goal (located in a room), a door, and a key (see Figure \ref{fig:anchor}). The robot is presented with a reward if it reaches the goal. Using this sparse reward signal, it must learn to first navigate to the key, use this to open the door, and then navigate to the goal. In this setting, an environment corresponds to a particular configuration of the key, door, goal, and obstacles. Different domains correspond to different key colors. 

We implement our approach on $5 \times 5$ grid-worlds using MiniGrid \cite{Boisvert18}. Observations in MiniGrid correspond to $5 \times 5 \times 3$ values; the three channels encode the object type (e.g, door), object color, and object state (e.g., open/closed) for a $5 \times 5$ neighborhood around the robot. 
The robot receives a sparse reward of $1-0.9t/T$, where $t$ is the time taken to reach the goal and $T = 250$ is the time-limit for completing the task. During training, the robot has access to environments from two domains corresponding to red and green keys. We use 48 training environments split evenly between these domains. At test-time, the robot is placed in environments with grey keys. This color choice is motivated by the fact that in MiniGrid, colors are encoded using integers (e.g., red: 0, green: 1), and grey corresponds to the color that is ``furthest away" in terms of this encoding (grey: 5). 
% For any given environment, the color of the key and the color of door are the same. This ensures that problem is always feasible, i.e. the robot will always be able to reach the goal if it learns the optimal policy. 
We implement IPO with an actor-critic architecture and the Fixed-$\Phi$ option (given the results in Section \ref{sec:lqr} and the arguments in Appendix \ref{sec:phi}); details and hyperparameters are provided in Appendix \ref{sec:colored keys hyperparams}.
Table \ref{tab:colored keys} reports the average rewards on 50 test environments from the training and test domains. We compare our approach to PPO \cite{Schulman17} trained to maximize rewards combined across training environments. As the table illustrates, IPO achieves better generalization to the new domain and is also more consistent across training seeds. %\anoop{I think it might be interesting to add another table of results but with a lower number of frames for eg. 80K instead of the current 120K. I think this might help exemplify how much better IPO can perform even for low amounts of data which might be an interesting result for other researchers especially ones doing zero/few-shot generalization using minimal data. Perhaps a graph of training and test accuracy for both algorithms as a function of number of frames could be interesting too. I'm also concerned that since the grid is size $5\times5$, PPO might do well for larger number of frames by memorizing a series of actions and accidentally running into the key at test time.} \ani{I don't think we see any benefits for low number of frames. IPO takes a bit longer to converge than PPO since I found that we need a lower learning rate to get good training. As for the $5 \times 5$ size, the obstacle and key locations are different for different environments. So, I don't think a memorization strategy works. Also, for larger grid sizes, we need policies with memory. This makes the training more complicated.}%

\begin{table}[h!]
\footnotesize
\begin{center}
% \vspace{-7pt}
  \begin{tabular}{ | l | c | c | c | }
    \hline 
    Key color & Red (training) & Green (training) & Grey (testing) \\ \hline
    \hline
    PPO &  0.94$\pm$0.004  & 0.94$\pm$0.005 & 0.80$\pm$0.12 \\ \hline
    IPO (ours) &  0.94$\pm$0.003  & 0.94$\pm$0.003  & {\bf 0.85$\pm$0.03} \\ \hline
    %IPO-rewards &  0$\pm$0  & 0$\pm$0  & 0$\pm$0 \\
    % \hline
  \end{tabular}
  \vspace{-2pt}
    \caption[]{\footnotesize{Colored-key domains: comparison of the average reward on 50 test environments drawn from different domains. The reported mean and std. dev. are across 10 different seeds. }  \label{tab:colored keys}}
  \end{center}
  \vspace{-10pt}
  \end{table}

\begin{table}[h!]
\footnotesize
\begin{center}
\vspace{-10pt}
  \begin{tabular}{ | l | c | c | c | c | c | }
    \hline 
    Friction & 1.1 & 1.2 & 1.3 & 1.4 & 1.5 \\ \hline
    \hline
    PPO &  96.6$\pm$2.1  & 94.0$\pm$3.6 & 89.8$\pm$8.1 & 83.4$\pm$12.0 & 78.2$\pm$13.7 \\ \hline
    IPO (ours) &  {\bf 99.0$\pm$1.3}  & {\bf 96.0$\pm$2.7}  & {\bf 93.8$\pm$5.9} & {\bf 87.4$\pm$9.8} & {\bf 81.0$\pm$12.6} \\ \hline
  \end{tabular}
 \vspace{-2pt}
\caption[]{\footnotesize{Door-opening environments. Training was performed using two domains with 0.0 and 0.1 friction respectively. The table compares the door-opening success rates for PPO and IPO on 100 test environments with higher friction. The reported mean and std. dev. are across 5 seeds.}  \label{tab:doorgym}}
  \end{center}
  \vspace{-20pt}
  \end{table}
  
\subsection{DoorGym}
\label{sec:doorgym}

Finally, we consider a more challenging task where a robot must learn to open doors (Figure \ref{fig:doorgym} (a)). We implement this example using Doorgym \cite{Urakami19}, which uses MuJoCo \cite{Todorov12} as its simulation engine. Each environment in Doorgym corresponds to a particular setting of physical parameters including door position, height, width, mass, handle location, door hinge spring constant, and hinge friction. Observations correspond to the pose (and velocity) of the gripper (hook), along with the position of the door handle. Actions correspond to torques applied on the gripper. During training, we use environments from two domains. 
The first domain corresponds to environments with \emph{zero} friction at the door hinge, while the second domain corresponds to the friction set to 0.1; all other environment parameters (e.g., door position, height, etc.) are randomized. We use 32 training environments, with a 75\%-25\% split between the two domains.  

We implement IPO using policies with an actor-critic architecture and the Fixed-$\Phi$ option. We compare IPO with PPO implemented using the same architecture and trained using environments pooled from both domains. Details on hyperparameters are reported in Appendix \ref{sec:doorgym hyperparams}. We use the reward function specified in \cite{Urakami19} for training. %, but do not perform domain randomization (i.e., training environments are fixed across training epochs). 
Table \ref{tab:doorgym} presents the door-opening success rates of both methods on test environments with the friction set to higher values than seen during training. As the table illustrates, IPO demonstrates improved generalization to these test domains. Perhaps more interesting than the quantitative difference between the two methods, we observe a \emph{qualitative} difference between the policies learned using IPO and PPO. Across five seeds, PPO learns a policy that uses the \emph{outside of the hook} on three seeds (Figure \ref{fig:doorgym} (b)). On one seed, PPO learns to use the hook as one would expect (Figure \ref{fig:doorgym} (c)) and demonstrates a mixture of the two behaviors on the other seed. While using the outside of the hook is a viable strategy for lower friction values, this strategy is not as robust for higher values. In contrast, IPO learns to perform the hooking maneuver shown in Figure \ref{fig:doorgym} (c) on all five seeds. A video of both policies is available at \href{https://youtu.be/J3O474yZ2Tc}{https://youtu.be/J3O474yZ2Tc}.

\section{Discussion and Conclusions}
\label{sec:conclusions}

% We have presented Invariant Policy Optimization (IPO) to tackle the challenging problem of learning policies with strong generalization beyond training domains in RL settings. 

We have considered the problem of learning policies with strong generalization beyond training domains. The key idea behind Invariant Policy Optimization (IPO) is to learn representations that support \emph{invariant action prediction} across different domains. We implemented the proposed techniques on: (i) linear quadratic regulator (LQR) problems with distractor observations, (ii) an example where an agent must learn to navigate to a goal by opening a door using different colored keys in its environment, and (iii) an example where a robot must learn to open doors with varying physical properties. We compared our approach with standard policy gradient methods (e.g., PPO) and demonstrated significant improvements in generalization performance on unseen domains. 

{\bf Future work.} On the theoretical front, an important direction for future work is to provide rigorous guarantees on generalization to novel domains. One potential avenue is to combine the algorithmic techniques presented here with recent results on PAC-Bayes generalization theory applied to control and RL settings \cite{Majumdar20, Veer20}. On the algorithmic front, an interesting direction is to use domain randomization techniques to \emph{automatically generate} new training domains that can be used to improve invariant policy learning (e.g., automatically generating domains with different colored keys in the colored-keys example). Finally, a particularly promising future direction is to explore the application of IPO to robotics problems involving sim-to-real transfer, where one considers simulation and reality as different domains to learn a policy that is invariant across them. % such that the "quirks" of the simulator get discarded. 

% We believe that the approaches presented here and the indicated future directions hold the promise of significantly improving the generalization performance of robotic systems trained using reinforcement learning approaches. 

% Acknowledgments---Will not appear in anonymized version
\section*{Acknowledgements}
\begin{small}
The authors are grateful to Kartik Ahuja for helpful clarifications on the training procedure for IRM-Games. The authors would like to thank Richard Song and Behnam Neyshabur for providing access to their code from \cite{Song19} for the LQR example in Section \ref{sec:lqr}. The authors are also grateful to Damien Teney for helpful comments on an early version of this paper.
\end{small}

\section*{Funding}
\begin{small}
This work is partially supported by the Office of Naval Research [Award Number:
N00014-18-1-2873], the Google Faculty Research Award, the
Amazon Research Award, the National Science Foundation [IIS-1755038], and the Alfred Rheinstein Faculty Award from Princeton University.
\end{small}

\bibliographystyle{abbrvnat} 
\bibliography{irom.bib}

\begin{thebibliography}{39}
\providecommand{\natexlab}[1]{#1}
\providecommand{\url}[1]{\texttt{#1}}
\expandafter\ifx\csname urlstyle\endcsname\relax
  \providecommand{\doi}[1]{doi: #1}\else
  \providecommand{\doi}{doi: \begingroup \urlstyle{rm}\Url}\fi

\bibitem[Agarwal et~al.(2019)Agarwal, Bullins, Hazan, Kakade, and
  Singh]{Agarwal19}
N.~Agarwal, B.~Bullins, E.~Hazan, S.~M. Kakade, and K.~Singh.
\newblock Online control with adversarial disturbances.
\newblock \emph{arXiv preprint arXiv:1902.08721}, 2019.

\bibitem[Ahuja et~al.(2020)Ahuja, Shanmugam, Varshney, and Dhurandhar]{Ahuja20}
K.~Ahuja, K.~Shanmugam, K.~Varshney, and A.~Dhurandhar.
\newblock Invariant risk minimization games.
\newblock \emph{arXiv preprint arXiv:2002.04692}, 2020.

\bibitem[Akkaya et~al.(2019)Akkaya, Andrychowicz, Chociej, Litwin, McGrew,
  Petron, Paino, Plappert, Powell, Ribas, et~al.]{Akkaya19}
I.~Akkaya, M.~Andrychowicz, M.~Chociej, M.~Litwin, B.~McGrew, A.~Petron,
  A.~Paino, M.~Plappert, G.~Powell, R.~Ribas, et~al.
\newblock Solving {R}ubik's cube with a robot hand.
\newblock \emph{arXiv preprint arXiv:1910.07113}, 2019.

\bibitem[Anderson and Moore(2007)]{Anderson07}
B.~Anderson and J.~Moore.
\newblock \emph{Optimal Control: Linear Quadratic Methods}.
\newblock Courier Corporation, 2007.

\bibitem[Arjovsky et~al.(2019)Arjovsky, Bottou, Gulrajani, and
  Lopez-Paz]{Arjovsky19}
M.~Arjovsky, L.~Bottou, I.~Gulrajani, and D.~Lopez-Paz.
\newblock Invariant risk minimization.
\newblock \emph{arXiv preprint arXiv:1907.02893}, 2019.

\bibitem[Bareinboim et~al.(2012)Bareinboim, Brito, and Pearl]{Bareinboim12}
E.~Bareinboim, C.~Brito, and J.~Pearl.
\newblock Local characterizations of causal bayesian networks.
\newblock In \emph{Graph Structures for Knowledge Representation and
  Reasoning}, pages 1--17. Springer, 2012.

\bibitem[Barron et~al.(2010)Barron, Goebel, and Jensen]{Barron10}
E.~Barron, R.~Goebel, and R.~Jensen.
\newblock Best response dynamics for continuous games.
\newblock \emph{Proceedings of the American Mathematical Society}, 138\penalty0
  (3):\penalty0 1069--1083, 2010.

\bibitem[Chevalier-Boisvert et~al.(2018)Chevalier-Boisvert, Willems, and
  Pal]{Boisvert18}
M.~Chevalier-Boisvert, L.~Willems, and S.~Pal.
\newblock Minimalistic gridworld environment for openai gym.
\newblock \url{https://github.com/maximecb/gym-minigrid}, 2018.

\bibitem[Cobbe et~al.(2019{\natexlab{a}})Cobbe, Hesse, Hilton, and
  Schulman]{Cobbe19b}
K.~Cobbe, C.~Hesse, J.~Hilton, and J.~Schulman.
\newblock Leveraging procedural generation to benchmark reinforcement learning.
\newblock \emph{arXiv preprint arXiv:1912.01588}, 2019{\natexlab{a}}.

\bibitem[Cobbe et~al.(2019{\natexlab{b}})Cobbe, Klimov, Hesse, Kim, and
  Schulman]{Cobbe19a}
K.~Cobbe, O.~Klimov, C.~Hesse, T.~Kim, and J.~Schulman.
\newblock Quantifying generalization in reinforcement learning.
\newblock In \emph{Proceedings of the International Conference on Machine
  Learning}, pages 1282--1289, 2019{\natexlab{b}}.

\bibitem[Dasgupta et~al.(2019)Dasgupta, Wang, Chiappa, Mitrovic, Ortega,
  Raposo, Hughes, Battaglia, Botvinick, and Kurth-Nelson]{Dasgupta19}
I.~Dasgupta, J.~Wang, S.~Chiappa, J.~Mitrovic, P.~Ortega, D.~Raposo, E.~Hughes,
  P.~Battaglia, M.~Botvinick, and Z.~Kurth-Nelson.
\newblock Causal reasoning from meta-reinforcement learning.
\newblock \emph{arXiv preprint arXiv:1901.08162}, 2019.

\bibitem[Dean et~al.(2019)Dean, Mania, Matni, Recht, and Tu]{Dean19}
S.~Dean, H.~Mania, N.~Matni, B.~Recht, and S.~Tu.
\newblock On the sample complexity of the linear quadratic regulator.
\newblock \emph{Foundations of Computational Mathematics}, pages 1--47, 2019.

\bibitem[Fazel et~al.(2018)Fazel, Ge, Kakade, and Mesbahi]{Fazel18}
M.~Fazel, R.~Ge, S.~M. Kakade, and M.~Mesbahi.
\newblock Global convergence of policy gradient methods for the linear
  quadratic regulator.
\newblock \emph{arXiv preprint arXiv:1801.05039}, 2018.

\bibitem[Goyal et~al.(2019)Goyal, Islam, Strouse, Ahmed, Botvinick, Larochelle,
  Bengio, and Levine]{Goyal19}
A.~Goyal, R.~Islam, D.~Strouse, Z.~Ahmed, M.~Botvinick, H.~Larochelle,
  Y.~Bengio, and S.~Levine.
\newblock Infobot: Transfer and exploration via the information bottleneck.
\newblock \emph{arXiv preprint arXiv:1901.10902}, 2019.

\bibitem[Ioffe and Szegedy(2015)]{Ioffe15}
S.~Ioffe and C.~Szegedy.
\newblock Batch normalization: Accelerating deep network training by reducing
  internal covariate shift.
\newblock \emph{arXiv preprint arXiv:1502.03167}, 2015.

\bibitem[Kingma and Ba(2014)]{Kingma14}
D.~P. Kingma and J.~Ba.
\newblock Adam: A method for stochastic optimization.
\newblock \emph{arXiv preprint arXiv:1412.6980}, 2014.

\bibitem[Krueger et~al.(2020)Krueger, Caballero, Jacobsen, Zhang, Binas, Priol,
  and Courville]{Krueger20}
D.~Krueger, E.~Caballero, J.-H. Jacobsen, A.~Zhang, J.~Binas, R.~L. Priol, and
  A.~Courville.
\newblock Out-of-distribution generalization via risk extrapolation {(REx)}.
\newblock \emph{arXiv preprint arXiv:2003.00688}, 2020.

\bibitem[Majumdar et~al.(2020)Majumdar, Farid, and Sonar]{Majumdar20}
A.~Majumdar, A.~Farid, and A.~Sonar.
\newblock {PAC}-{B}ayes {C}ontrol: Learning policies that provably generalize
  to novel environments.
\newblock \emph{International Journal of Robotics Research (IJRR)}, October
  2020.

\bibitem[Mnih et~al.(2013)Mnih, Kavukcuoglu, Silver, Graves, Antonoglou,
  Wierstra, and Riedmiller]{Mnih13}
V.~Mnih, K.~Kavukcuoglu, D.~Silver, A.~Graves, I.~Antonoglou, D.~Wierstra, and
  M.~Riedmiller.
\newblock Playing {A}tari with deep reinforcement learning.
\newblock \emph{arXiv preprint arXiv:1312.5602}, 2013.

\bibitem[Nair et~al.(2019)Nair, Zhu, Savarese, and Fei-Fei]{Nair19}
S.~Nair, Y.~Zhu, S.~Savarese, and L.~Fei-Fei.
\newblock Causal induction from visual observations for goal directed tasks.
\newblock \emph{arXiv preprint arXiv:1910.01751}, 2019.

\bibitem[Nichol et~al.(2018)Nichol, Pfau, Hesse, Klimov, and
  Schulman]{Nichol18}
A.~Nichol, V.~Pfau, C.~Hesse, O.~Klimov, and J.~Schulman.
\newblock Gotta learn fast: A new benchmark for generalization in reinforcement
  learning.
\newblock \emph{arXiv preprint arXiv:1804.03720}, 2018.

\bibitem[Pacelli and Majumdar(2020)]{Pacelli20}
V.~Pacelli and A.~Majumdar.
\newblock Learning task-driven control policies via information bottlenecks.
\newblock In \emph{Proceedings of Robotics: Science and Systems (RSS)}, 2020.

\bibitem[Pearl(2009)]{Pearl09}
J.~Pearl.
\newblock \emph{Causality}.
\newblock Cambridge University Press, 2009.

\bibitem[Peng et~al.(2018)Peng, Andrychowicz, Zaremba, and Abbeel]{Peng18}
X.~B. Peng, M.~Andrychowicz, W.~Zaremba, and P.~Abbeel.
\newblock Sim-to-real transfer of robotic control with dynamics randomization.
\newblock In \emph{IEEE International Conference on Robotics and Automation
  (ICRA)}. IEEE, 2018.

\bibitem[Peters et~al.(2016)Peters, B{\"u}hlmann, and Meinshausen]{Peters16}
J.~Peters, P.~B{\"u}hlmann, and N.~Meinshausen.
\newblock Causal inference by using invariant prediction: identification and
  confidence intervals.
\newblock \emph{Journal of the Royal Statistical Society: Series B (Statistical
  Methodology)}, 78\penalty0 (5):\penalty0 947--1012, 2016.

\bibitem[Peters et~al.(2017)Peters, Janzing, and Sch{\"o}lkopf]{Peters17}
J.~Peters, D.~Janzing, and B.~Sch{\"o}lkopf.
\newblock \emph{Elements of Causal Inference: Foundations and Learning
  Algorithms}.
\newblock MIT Press, 2017.

\bibitem[Schulman et~al.(2017)Schulman, Wolski, Dhariwal, Radford, and
  Klimov]{Schulman17}
J.~Schulman, F.~Wolski, P.~Dhariwal, A.~Radford, and O.~Klimov.
\newblock Proximal policy optimization algorithms.
\newblock \emph{arXiv preprint arXiv:1707.06347}, 2017.

\bibitem[Sinha et~al.(2017)Sinha, Namkoong, and Duchi]{Sinha17}
A.~Sinha, H.~Namkoong, and J.~Duchi.
\newblock Certifying some distributional robustness with principled adversarial
  training.
\newblock \emph{arXiv preprint arXiv:1710.10571}, 2017.

\bibitem[Song et~al.(2019)Song, Jiang, Du, and Neyshabur]{Song19}
X.~Song, Y.~Jiang, Y.~Du, and B.~Neyshabur.
\newblock Observational overfitting in reinforcement learning.
\newblock \emph{arXiv preprint arXiv:1912.02975}, 2019.

\bibitem[Srivastava et~al.(2014)Srivastava, Hinton, Krizhevsky, Sutskever, and
  Salakhutdinov]{Srivastava14}
N.~Srivastava, G.~Hinton, A.~Krizhevsky, I.~Sutskever, and R.~Salakhutdinov.
\newblock Dropout: a simple way to prevent neural networks from overfitting.
\newblock \emph{The Journal of Machine Learning Research}, 15\penalty0
  (1):\penalty0 1929--1958, 2014.

\bibitem[Tan et~al.(2018)Tan, Zhang, Coumans, Iscen, Bai, Hafner, Bohez, and
  Vanhoucke]{Tan18}
J.~Tan, T.~Zhang, E.~Coumans, A.~Iscen, Y.~Bai, D.~Hafner, S.~Bohez, and
  V.~Vanhoucke.
\newblock Sim-to-real: Learning agile locomotion for quadruped robots.
\newblock \emph{arXiv preprint arXiv:1804.10332}, 2018.

\bibitem[Teney et~al.(2020)Teney, Abbasnejad, and Hengel]{Teney20}
D.~Teney, E.~Abbasnejad, and A.~v.~d. Hengel.
\newblock Unshuffling data for improved generalization.
\newblock \emph{arXiv preprint arXiv:2002.11894}, 2020.

\bibitem[Thrun et~al.(2005)Thrun, Burgard, and Fox]{Thrun05}
S.~Thrun, W.~Burgard, and D.~Fox.
\newblock \emph{Probabilistic Robotics}.
\newblock MIT Press, 2005.

\bibitem[Tobin et~al.(2017)Tobin, Zaremba, and Abbeel]{Tobin17}
J.~Tobin, W.~Zaremba, and P.~Abbeel.
\newblock Domain randomization and generative models for robotic grasping.
\newblock \emph{arXiv preprint arXiv:1710.06425}, 2017.

\bibitem[Todorov et~al.(2012)Todorov, Erez, and Tassa]{Todorov12}
E.~Todorov, T.~Erez, and Y.~Tassa.
\newblock Mujoco: A physics engine for model-based control.
\newblock In \emph{Proceedings of the IEEE/RSJ International Conference on
  Intelligent Robots and Systems (IROS)}, pages 5026--5033. IEEE, 2012.

\bibitem[Tu and Recht(2018)]{Tu18}
S.~Tu and B.~Recht.
\newblock The gap between model-based and model-free methods on the linear
  quadratic regulator: An asymptotic viewpoint.
\newblock \emph{arXiv preprint arXiv:1812.03565}, 2018.

\bibitem[Urakami et~al.(2019)Urakami, Hodgkinson, Carlin, Leu, Rigazio, and
  Abbeel]{Urakami19}
Y.~Urakami, A.~Hodgkinson, C.~Carlin, R.~Leu, L.~Rigazio, and P.~Abbeel.
\newblock Doorgym: A scalable door opening environment and baseline agent.
\newblock \emph{arXiv preprint arXiv:1908.01887}, 2019.

\bibitem[Veer and Majumdar(2020)]{Veer20}
S.~Veer and A.~Majumdar.
\newblock Probably approximately correct vision-based planning using motion
  primitives.
\newblock In \emph{Proceedings of the Conference on Robot Learning (CoRL)},
  2020.

\bibitem[Zhang et~al.(2020)Zhang, Lyle, Sodhani, Filos, Kwiatkowska, Pineau,
  Gal, and Precup]{Zhang20}
A.~Zhang, C.~Lyle, S.~Sodhani, A.~Filos, M.~Kwiatkowska, J.~Pineau, Y.~Gal, and
  D.~Precup.
\newblock Invariant causal prediction for block {MDP}s.
\newblock \emph{arXiv preprint arXiv:2003.06016}, 2020.

\end{thebibliography}

\appendix

% \newpage
\section{Appendix}
\label{sec:appendix}

\subsection{Details on policy averaging}
\label{sec:averaging}

For stochastic policies on discrete action spaces with cardinality $n_a$ (e.g., the colored-key example), we define the averaged policy $\pi^\text{av}$ as follows. Each domain $d$ has an associated $\hat{\pi}^d$ that assigns a score vector $\hat{\pi}^d(h_t) \in \mathbb{R}^{n_a}$ to actions for a given $h_t$ (recall that $h_t = \Phi(o_t)$, where $o_t$ is an observation). We define $\hat{\pi}^\text{av}(h_t) := \frac{1}{n_d} \sum_{i=1}^{n_d} \hat{\pi}^d(h_t)$. The probabilities assigned to each action by the averaged policy $\pi^\text{av}$ are then calculated by passing the score vector $\hat{\pi}^\text{av}(h_t)$ through a softmax function. 

For continuous action spaces (e.g., the DoorGym example in Section \ref{sec:doorgym}), we parameterize distributions over the action space using Gaussians (with diagonal covariance matrices). Let $n_a$ denote the dimension of the action space. For a given $h_t$, we output mean $\mu_t^d(h_t) \in \mathbb{R}^{n_a}$ and standard deviation $\sigma_t^d(h_t) \in \mathbb{R}^{n_a}$ for each domain $d$. The Gaussian distribution over actions corresponding to policy $\pi^\text{av}$ is then defined by mean $\frac{1}{n_d} \sum_{i=1}^{n_d} \mu_t^d(h_t)$ and standard deviation $\frac{1}{n_d} \sum_{i=1}^{n_d} \sigma_t^d(h_t)$. 

% v, q

% We endow each domain $d$ with its own policy $\pi^d$ and define an overall ensemble policy $\pi^\text{av}(\cdot | h_t)$ by $\frac{1}{n_d} \sum_{d=1}^{n_d} \pi^d(\cdot | h_t)$.

\subsection{Discussion on choice of $\Phi$ = I}
\label{sec:phi}

Fixing $\Phi$ = I (identity) is an effective approach and can (under certain conditions) recover invariant predictors that extract causal features and ignore non-causal factors. The intuition is that the burden of extracting causal features from observations is simply shifted to $\pi$ (the portion of the overall policy that maps the output of $\Phi$ to actions). Concretely, consider a setting where the observation $o_t = [o_\text{c}, o_\text{nc}]$ consists of two components corresponding to causal features $o_\text{c}$ and non-causal features $o_\text{nc}$ (e.g., color). One can construct an invariant policy of the form $o_t \mapsto \Phi(o_t) \mapsto \pi(\Phi(o_t))$, where $\Phi$ extracts the causal features from the observation (i.e., $\Phi$ maps $o_t$ to $o_\text{c}$). Now, suppose one had chosen $\Phi$ = I for this example. Then, one can still construct an invariant policy: $o_t \mapsto \text{I}(o_t) \mapsto \pi'(\text{I}(o_t))$, where $\pi'$ is $\pi(\Phi(\cdot))$. Hence, the overall policy is still extracting causal features, but the burden of this extraction has simply been transferred to $\pi'$. Hence the IPO best-response training should, in principle, still be able to find an invariant policy with $\Phi$ = I. The paper \cite{Ahuja20} provides a more formal characterization of conditions under which one may choose $\Phi$ = I.

\subsection{Computing platform}

The examples presented in Section \ref{sec:examples} are implemented on a desktop computer with six 3.50GHz Intel i7-7800X processors, 32GB RAM, and four Nvidia GeForce RTX 2080 GPUs. 

\subsection{Hyperparameters for LQR example}
\label{sec:lqr hyperparams}

We use the Adam optimizer \cite{Kingma14} for our experiments with the learning rates shown in Table \ref{tab:lqr hyperparams}. 

\begin{table}[h!]
\footnotesize
\begin{center}
  \begin{tabular}{ | l | c |}
    \hline 
     & Learning rate  \\ \hline
    \hline
    Gradient descent &  0.001   \\ \hline
    Overparameterization &  0.001 \\ \hline
    IPO &  0.0005  \\ \hline
    %IPO-rewards (exploitation) &  0.001  \\
    % \hline
  \end{tabular}
  \vspace{5pt}
    \caption[]{\footnotesize{Learning rates for LQR problem.}  \label{tab:lqr hyperparams}}
  \end{center}
  \vspace{-10pt}
  \end{table}

\subsection{Hyperparameters for the Colored-Key example}
\label{sec:colored keys hyperparams}

\begin{figure}[h]
\begin{center}
\includegraphics[width=1.0\columnwidth]{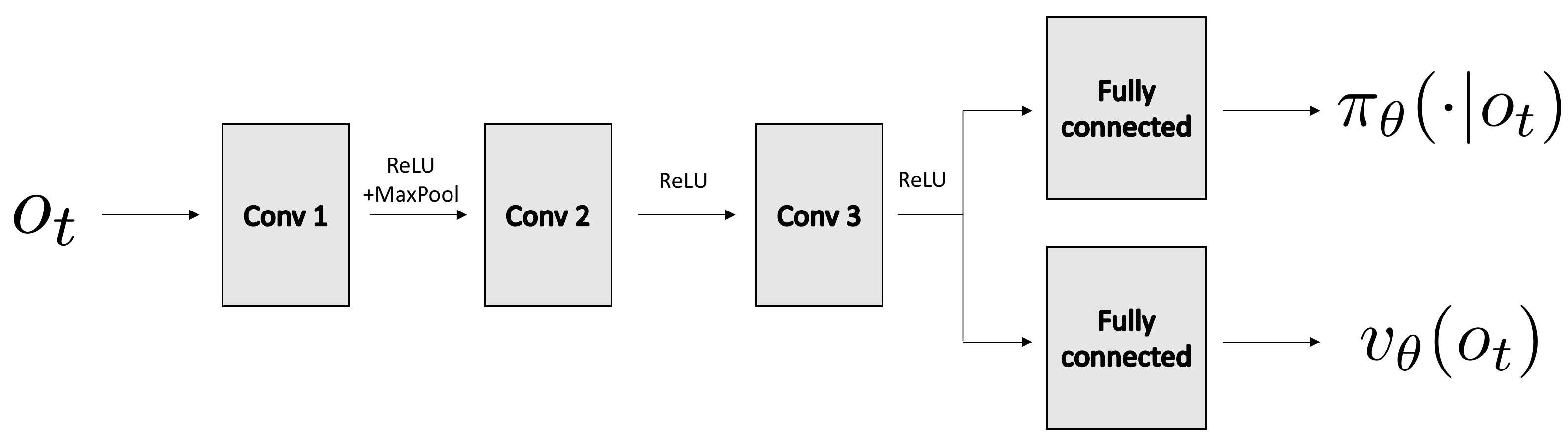}
\end{center} 
% \vspace{-14pt}
\caption{\footnotesize{Actor-critic architecture used for colored-keys example.} \label{fig:colored keys architecture}}
% \vspace{-10pt}
\end{figure}

We use the default actor-critic architecture (with no memory) used to train agents using PPO in MiniGrid \cite{Boisvert18}. This is shown in Figure \ref{fig:colored keys architecture}. The hyperparameters for PPO are also the default ones used for MiniGrid (see Table \ref{tab:colored keys hyperparams}). For IPO, the policy associated with each domain utilizes the same architecture shown in Figure \ref{fig:colored keys architecture}. We implement IPO using the Fixed-$\Phi$ option. The parameters used for the policy-update step in IPO are shown in Table \ref{tab:colored keys hyperparams}. These are identical to the ones used for PPO, with the exception of a lower learning rate. 

\begin{table}[h!]
\footnotesize
\begin{center}
  \begin{tabular}{ | c | c | c |}
    \hline 
     & PPO & IPO  \\ \hline
    \hline
    \# time-steps per rollout on environment &  128 & 128   \\ \hline
    Epochs per rollout &  4 & 4 \\ \hline
    Discount $\gamma$ &  0.99 & 0.99  \\ \hline
    GAE $\lambda$ &  0.95 & 0.95  \\ \hline
    Batch size &  256 & 256  \\ \hline
    Entropy bonus &  0.01 & 0.01  \\ \hline
    PPO clip range &  0.2 & 0.2  \\ \hline
    Learning rate &  0.001 & 0.0005  \\ \hline
    Total time-steps &  120K & 120K  \\ \hline
  \end{tabular}
  \vspace{5pt}
    \caption[]{\footnotesize{Hyperparameters for colored-keys example.}  \label{tab:colored keys hyperparams}}
  \end{center}
  \vspace{-10pt}
  \end{table}

\subsection{Hyperparameters for the DoorGym example}
\label{sec:doorgym hyperparams}

We use the default policy network architecture from DoorGym \cite{Urakami19}. The hyperparameters for PPO are also the default ones (see Table \ref{tab:doorgym hyperparams}). For IPO, the policy corresponding to each domain is the same as PPO. We implement IPO using the Fixed-$\Phi$ option. The parameters used for the policy-update step in IPO are shown in Table \ref{tab:doorgym hyperparams}. These are identical to the ones used for PPO, with the exception of a lower learning rate.

\begin{table}[h!]
\footnotesize
\begin{center}
  \begin{tabular}{ | c | c | c |}
    \hline 
     & PPO & IPO  \\ \hline
    \hline
    \# time-steps per rollout on environment &  4096 & 4096   \\ \hline
    Epochs per rollout &  10 & 10 \\ \hline
    Discount $\gamma$ &  0.99 & 0.99  \\ \hline
    GAE $\lambda$ &  0.95 & 0.95  \\ \hline
    Batch size &  256 & 256  \\ \hline
    Entropy bonus &  0.01 & 0.01  \\ \hline
    PPO clip range &  0.2 & 0.2  \\ \hline
    Learning rate &  0.001 & 0.0005  \\ \hline
    Total time-steps &  25M & 25M  \\ \hline
  \end{tabular}
  \vspace{5pt}
    \caption[]{\footnotesize{Hyperparameters for Doorgym example.}  \label{tab:doorgym hyperparams}}
  \end{center}
  \vspace{-10pt}
  \end{table}

\end{document}